\documentclass[runningheads]{llncs}
\usepackage{graphicx}
\usepackage{amsmath,amssymb} 

\usepackage{fancyhdr}

\usepackage{enumitem}
\usepackage{color}
\usepackage{hyperref}
\usepackage{cite}
\usepackage{url}
\usepackage{amsmath,amssymb}
\usepackage{mathtools}
\usepackage{graphicx}
\usepackage{wrapfig}
\usepackage{xcolor}
\usepackage{tablefootnote}
\usepackage{threeparttable}
\newcommand{\round}[1]{\ensuremath{\lfloor#1\rceil}}
\usepackage[ruled,vlined]{algorithm2e}
\SetKwRepeat{Do}{do}{while}

\begin{document}



\newcommand{\ind}{\mathbb{I}}

\newcommand{\drel}{D^+_{\mathrm{rel}}}

\newcommand{\co}{\omega}

\newcommand{\Co}{\widetilde{C}^\omega}
\pagestyle{headings}
\mainmatter
\def\ECCV16SubNumber{5083}  

\title{Weight Fixing Networks }

\titlerunning{Weight Fixing Networks}

\authorrunning{Subia-Waud et al.}

\author{Christopher Subia-Waud \& Srinandan Dasmahapatra}
\institute{University of Southampton, Southampton, SO17 1BJ \\ \email{\{cc2u18, sd\}@soton.ac.uk}}

\maketitle

\begin{abstract}
Modern iterations of deep learning models contain millions (billions) of unique parameters—each represented by a $b$-bit number. Popular attempts at compressing neural networks (such as pruning and quantisation) have shown that many of the parameters are superfluous, which we can remove (pruning) or express with $b' < b$ bits (quantisation) without hindering performance.  Here we look to go much further in minimising the information content of networks. Rather than a channel or layer-wise encoding, we look to lossless whole-network quantisation to minimise the entropy and number of unique parameters in a network. We propose a new method,  which we call Weight Fixing Networks (WFN) that we design to realise four model outcome objectives: i) very few unique weights, ii) low-entropy weight encodings,  iii) unique weight values which are amenable to energy-saving versions of hardware multiplication, and iv)  lossless task-performance. Some of these goals are conflicting. To best balance these conflicts, we combine a few novel (and some well-trodden) tricks; a novel regularisation term, (i, ii) a view of clustering cost as relative distance change (i, ii, iv), and a focus on whole-network re-use of weights (i, iii). Our Imagenet experiments demonstrate lossless compression using 56x fewer unique weights and a 1.9x  lower weight-space entropy than SOTA quantisation approaches.  Code and model saves can be found at \href{https://github.com/subiawaud/Weight_Fix_Networks}{github.com/subiawaud/Weight\_Fix\_Networks}\footnote{Paper Published: 978-3-031-20082-3, ECCV 2022, Part XI, LNCS 13671}.

\end{abstract}

\section{Introduction}

Deep learning models have a seemingly inexorable trajectory toward growth. Growth in applicability,  performance,  investment, and optimism. Unfortunately, one area of growth is lamentable - the ever-growing energy and storage costs required to train and make predictions. To fully realise the promise of deep learning methods, work is needed to reduce these costs without hindering task performance.  

Here we look to contribute a methodology and refocus towards the goal of reducing both the number of bits to describe a network as well the total number of \emph{unique} weights in a network. The motivation to do so is driven both by practical considerations of accelerator designs \cite{Moons2016,  Han2016}, as well as the more theoretical persuasions of the \emph{Minimal Description Length} (MDL)  principle \cite{Rissanen1978, Nannen2010,  Barron1998} as a way of determining a \emph{good} model.

\begin{figure}
\center
\includegraphics[width= 0.8 \textwidth]{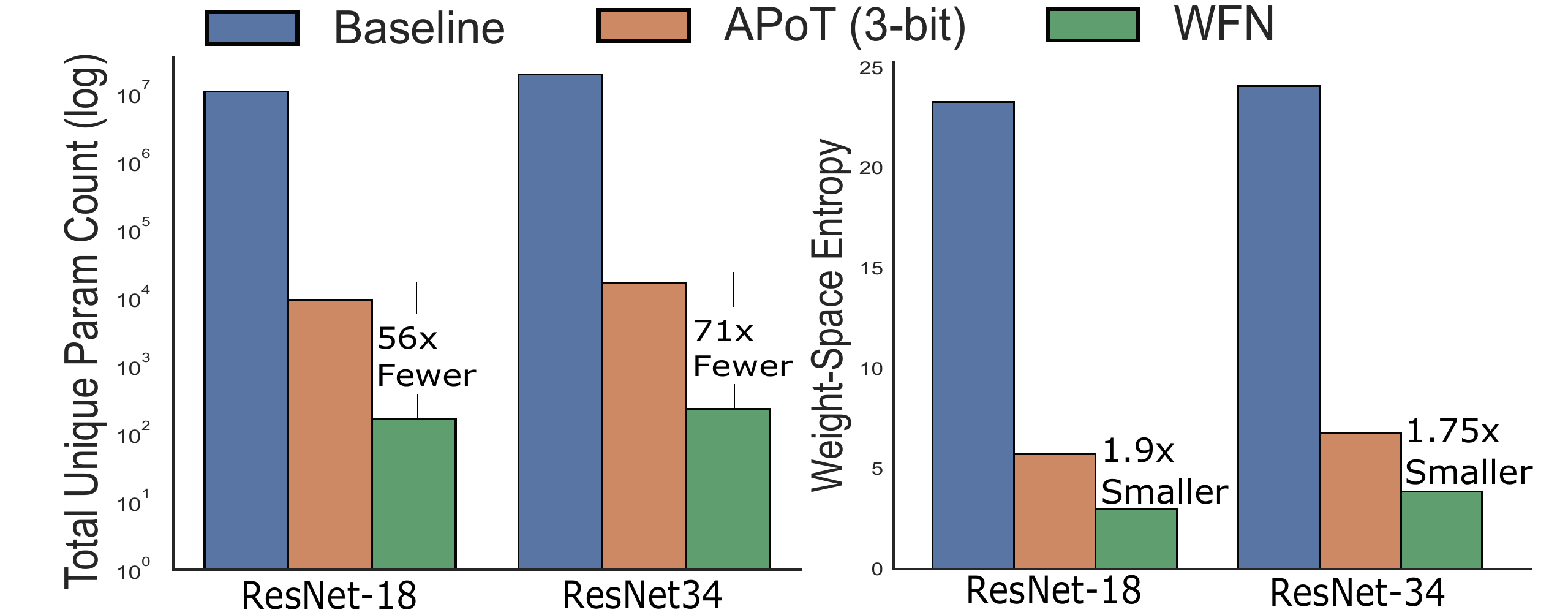}
\caption{WFN reduces the total number of weights and the entropy of a network far further than other quantisation works.  \textbf{Left}: The total number of unique parameters left after quantisation is 56x fewer than APoT for ResNet-18 trained on the Imagenet dataset and 71x for the ResNet-34 model.   \textbf{Right}: The entropy of the parameters across the network is 1.9x and 1.65x smaller when using the WFN approach over APoT. }
\end{figure}

\noindent \textbf{The Minimal Description Length.} Chaitin's hypothesis captures the thinking behind the MDL principle with the statement ``comprehension is compression" \cite{Blier2018, Chaitin2007}. That is, to learn is to compress. In this setting, the \emph{best} model minimises the combined cost of describing both the model and the prediction errors. Deep learning models have shown themselves adept at minimising the latter, but it is not clear that we sufficiently compress the model description through unbounded standard gradient descent training. One way to think about MDL in relation to deep learning compression is the following \footnote{originally posed in `Keeping neural networks simple by minimizing the description length of the weights' \cite{Hinton1993}}: 

Imagine that Bob wants to communicate ground-truth targets to Alice. To achieve this, he can forward both a predictive model and its errors, compressed to as few bits as possible without losing any information. Given these two components, Alice can pass the input data into the model and, given the communicated errors from Bob, make any adjustments to its output to retrieve the desired ground truth. This formulation is the \emph{two-part compression} approach to the problem of learning \cite{Blier2018}. The MDL principle says that the \emph{best} model is obtained by minimising the sum of the lengths of the codes for model and errors.


Although the MDL treatment of learning is largely theoretical, it has motivated the design of compressed network architectures  \cite{Hinton1993,  Havasi2018,  Hinton1996}. We believe that a more direct optimisation to minimise the  information theoretic content could bear fruit for downstream hardware translations, but let us start by setting out the description quantities we wish to minimise. 

\noindent \textbf{Describing a Solution.} Describing a classification model's error is well captured with the cross-entropy loss. From an information-theoretic perspective, the cross-entropy loss measures the average message length per data point needed to describe the difference in predicted and target output distributions. Essentially, large errors cost more to describe than small errors. 

The cost of describing a model is more complex, requiring two elements to be communicated -- the weight values, and their arrangement.  A metric used to capture both components is the \emph{representation cost}, as outlined in Deep K-means \cite{Wu2018}  (Equation \ref{eq:repcost} below). Here, the cost of describing the model is defined as the summed bit-width of each weight representation times the number of times each weight is used in an inference calculation.

\noindent \textbf{Minimising the Representation Costs in Accelerators.} Importantly, this representation cost as a model description can be translated directly into accelerator design savings, as shown by the seminal work in Deep Compression \cite{Mao2016} and subsequent  accelerator design, EIE \cite{Han2016}. Here the authors cluster neural networks' weights and use Huffman encoding to represent/describe the network cheaply. From an information-theoretic perspective, the motivation for Huffman encoding is simple; this encoding scheme is likely to give us a compressed result closest to our underlying weight-space entropy. However, this work was not focused on the information content, so why was it advantageous to an accelerator design? For this, we need to look at where the computation costs are most concentrated. 

The most expensive energy costs in inference calculations lie in memory reads \cite{Horowitz2014, Gao2017}. For every off-chip DRAM data read, we pay the equivalent of over two hundred 32-bit multiplications in energy costs\footnote{45nm process} \cite{Horowitz2014}. This energy cost concentration has led to the pursuit of data movement minimisation schemes using accelerator dataflow mappings \cite{Han2016, Chen2017, Chen2015, Chen2020}. These mappings aim to store as much of the network as possible close to computation and maximise weight re-use. From an algorithmic perspective, this makes networks with low entropy content desirable. To make use of a low entropy weight-space, a dataflow mapping can store compressed codewords for each weight which, when decoded, point to the address of the full precision weight, which itself is stored in cheaper access close-to-compute memory. The idea is that the addition of the codeword storage and access costs plus the full weight value access costs can be much smaller than in the unquantised network \cite{Mao2016a,   Wu2018}. Several accelerator designs have successfully implemented such a scheme \cite{Moons2016,  Han2016}.

As a simple illustrative example, let us define a filter in a network post-quantisation with the values $[900, 104, 211, 104, 104, 104, 399, 211, 104]$. This network has an entropy of 1.65, meaning each weight can be represented, on average, using a minimum of 1.65 bits, compared to the  9.8bits ($\log$(900)) needed for an uncompressed version of the same network. Using Huffman encoding, we get close to this bound by encoding the network weights as $w\mapsto c(w)$ with:

\[
c(104) = 1, c(211) = 01, c(399) = 001, c(900) = 000
\]


\noindent and the complete filter can be represented as ``000101111001011", totalling just 15 bits, 1.66 bits on average per value. Each decoded codeword points a corresponding weight in the reduced set of unique weights required for computation. These unique weights (since there are very few of them) can be stored close to compute on memory units, processing element scratchpads  or SRAM cache depending on the hardware flavour \cite{banakar2001comparison, banakar2002scratchpad}, all of which have minimal (almost free) access costs. The storage and data movement cost of the encoded weights plus the close-to-compute weight access should be smaller than the storage and movement costs of directly representing the network with the weight values. This link -- between minimising the model description and reducing accelerator representational costs -- motivates our approach. 

\noindent \textbf{Objectives.} So we ask ourselves what we could do algorithmically to maximise the benefit of accelerator dataflows and minimise the description length. Since Huffman encoding is used extensively in accelerator designs, we focus on finding networks that reduce the network description when compressed using this scheme. To do this, we first focus on reducing the number of  \emph{unique} weights a network uses. Fewer \emph{unique} weights whilst fixing the network topology and the total number of parameters will mean that more weights are re-used more often. Further gains can be achieved if we can concentrate the distribution of weights around a handful of values, enabling frequently used weights to be stored cheaply, close to compute. Finally, we ask what the ideal values of these weights would be. From a computational perspective, not all multiplications are created equal. Powers-of-two, for example, can be implemented as simple bit-shifts. Mapping the weights used most to these values offers potential further optimisation in hardware. Putting these three requirements together: few unique weights; a low-entropy encoding with a distribution of weights highly concentrated around a tiny subset of values; and a focus on powers-of-two values for weights --- all motivated to both minimise the MDL as well as the computation costs in accelerator designs --- we present our contribution.

\noindent \textbf{Weight Fixing Networks.} Our work's overarching objective is to transform a network comprising many weights of any value (limited only by value precision) to one with the same number of weights but just a few unique values and concentrate the weights around an even smaller subset of weights.   Rather than selecting the unique weights a priori, we let the optimisation guide the process in an iterative \emph{cluster-then-train} approach. We cluster an ever-increasing subset of weights to one of a few cluster centroids in each iteration. We map the pre-trained network weights to these cluster centroids, which constitute a pool of unique weights. The training stage follows standard gradient descent optimisation to minimise performance loss with two key additions. Firstly, only an ever decreasing subset of the weights are \emph{\text{free}} to be updated. We also use a new regularisation term to penalise weights with large relative distances to their nearest clusters. We iteratively cluster subsets of weights to their nearest cluster centre, with the way we determine which subset to move a core component of our contribution.

\begin{figure}
\center
\includegraphics[width = 1\textwidth]{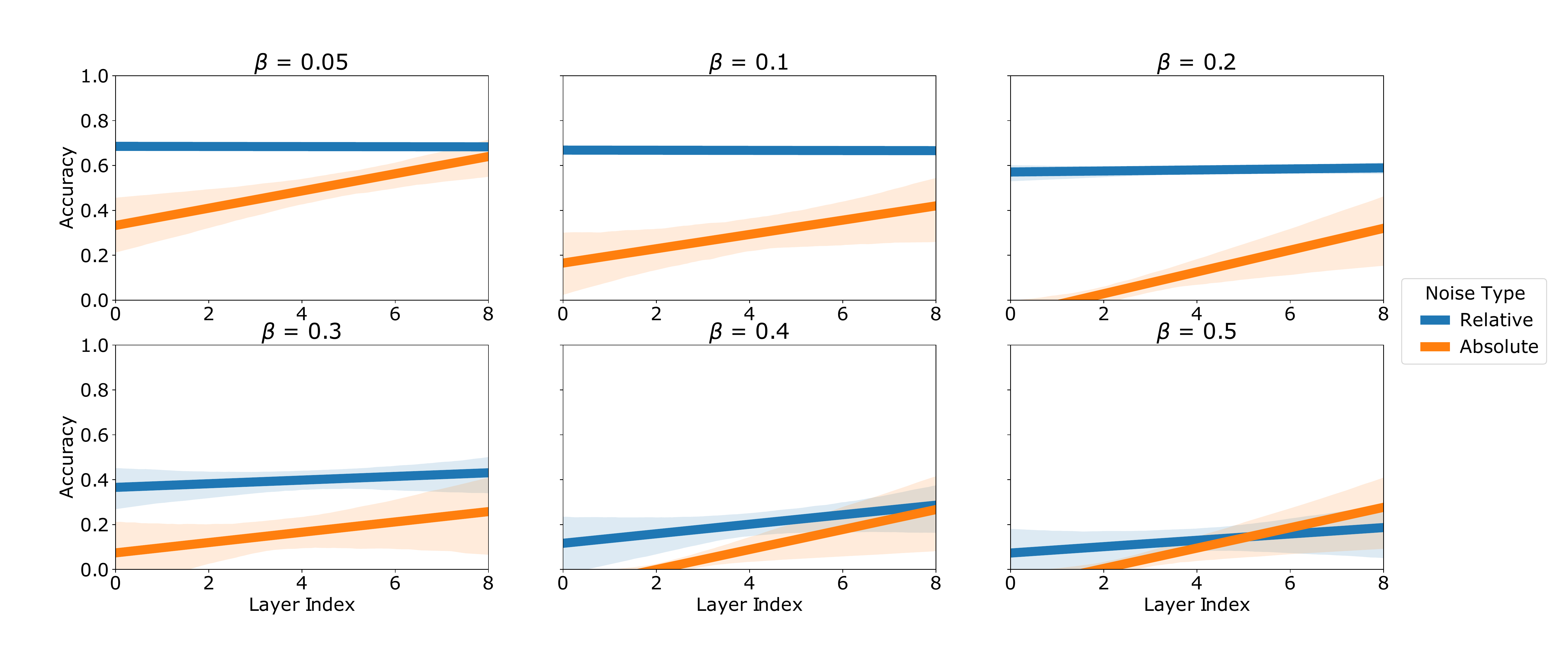}
\caption{We explore adding relative vs absolute noise to each of the layers (x-axis). The layer index indicates which layer was selected to have noise added. Each layer index is a separate experiment with the 95\% confidence intervals shaded.  }
 \label{fig:noise1}
\end{figure}

\noindent \textbf{Small Relative Distance Change.} Rather than selecting subsets with small Euclidean distances to cluster centres,  or those that have small magnitude \cite{Zhou2017},  we make the simple observation that the \emph{relative} -- as opposed to absolute -- weight change matters. We find that the tolerated distance  $\delta w_i$ we can move a weight  $w_i$ when quantised  depends on the relative distance  $|(\delta w_i/w_i)|$. When the new value  $w_i + \delta w_i = 0$  --- as is the case for pruning methods --- then the magnitude of the weight \emph{is} the distance. However, this is not the case more generally. We demonstrate the importance of quantising with small relative changes using simple empirical observations. Using a pre-trained ResNet-18 model, we measure changes to network accuracy when adding relative vs absolute noise to the layers' weights and measure the accuracy change. For relative noise  we choose a scale parameter $\beta|w_i^l|$ for each layer-$l$ weight $w_i^l$, and set $w_i^l \leftarrow w_i^l + \beta|w_i^l|\varepsilon$, $\varepsilon\sim \mathcal{N}(0,1)$. For additive noise perturbations, all weights $w_i^l$ are perturbed by the mean absolute value of weights $\overline{|w^l|}$ in layer $l$ scaled by $\beta$: $w_i^l\leftarrow w_i^l+\beta\overline{|w^l|}\varepsilon$.


We run each layer-$\beta$ combination experiment multiple times -- to account for fluctuation in the randomised noise -- and present the results in Figure \ref{fig:noise1}. Even though the mean variation of noise added is the same, noise relative to the original weight value (multiplicative noise) is much better tolerated than absolute (additive noise). Since moving weights to quantisation centres is analogous to adding noise, we translate these results into our approach and prioritise clustering weights with small relative distances first. We find that avoiding significant quantisation errors requires ensuring that $\frac{|\delta w_i|}{|w_i|}$ is small for all weights. If this is not possible, then performance could suffer. In this case, we create an additional cluster centroid in the vicinity of an existing cluster to reduce this relative distance. Our work also challenges the almost universal trend in the literature \cite{YuhangLiXinDong2020,  Jung2019,Zhang, Zhou2016, Yamamoto2021, oh2021}  of leaving the first and last layers either at full precision or 8-bit. Instead, we attempt a full network quantisation. The cost of not quantising the first layer -- which typically requires the most re-use of weights due to the larger resolution of input maps --  and the final linear layer -- which often contains the largest number of unique weight values -- is too significant to ignore.

With multiple stages of training and clustering, we finish with an appreciably reduced set of unique weights. We introduce a regularisation term that encourages non-uniform, high probability regions in the weight distribution to induce a lower-entropy weight-space.  The initial choice of cluster centroids as powers-of-two helps us meet our third objective -- energy-saving multiplication.  Overall we find four distinct advantages over the works reviewed:

\begin{itemize}
\item We assign a cluster value to \emph{all weights} --- including the first and last layers.
\item We emphasise a low entropy encoding with a regularisation term, achieving entropies smaller than those seen using  3-bit quantisation approaches -- over which we report superior performance. 
\item We require no additional layer-wise scaling, sharing the unique weights across all layers. 
\item WFN substantially reduces the number of unique parameters in a network when compared to existing SOTA quantisation approaches. 
\end{itemize}

%

\section{Related Work}



\noindent \textbf{Clip and Scale Quantisation.} Typical quantisation approaches reduce the number of bits used to represent components of the network.  Quantisation has been applied to all parts of the network with varying success; the weights, gradients, and activations have all been attempted \cite{Hubara2016, Lee2017, Jung2019, Yang2019, Shkolnik, Zhou2016}. Primarily, these approaches are motivated by the need to reduce the energy costs of the multiplication of 32-bit floating-point numbers.  This form of quantisation  maps the weight $w_i$ to $w_i'=s\, \mathrm{round}(w_i)$, where  $\mathrm{round}()$ is a predetermined rounding function and $s$ a scaling factor. The scaling factor (determined by a clipping range) can be learned channel-wise \cite{Jacob2018, Zhang},  or more commonly,  layerwise in separate formulations. This results in different channels/layers having a diverse pool of codebooks for the network weights/activations/gradients. Quantisation can be performed without training --- known as post-training quantisation, or with added training steps  -- called quantisation-aware training (QAT).  Retraining incurs higher initial costs but results in superior performance. 

A clipping+scaling quantisation example relevant to our own is the work of \cite{Zhou2017}, where the authors restrict the layerwise rounding of weights to powers-of-two. The use of powers-of-two has the additional benefit of energy-cheap bit-shift multiplication.  A follow-up piece of work \cite{YuhangLiXinDong2020} suggests additive powers-of-two (APoT) instead to capture the pre-quantised distribution of weights better.    \\

\noindent \textbf{Weight Sharing Quantisation.} Other formulations of quantisation do not use clipping and scaling factors \cite{Stock2019, Tartaglione2021, Wu2018}.  Instead, they adopt clustering techniques to cluster the weights and fix the weight values to their assigned group cluster centroid. These weights are stored as codebook indices, allowing for compressed representation methods such as Huffman encoding to squeeze the network further. 

 We build on these methods, which, unlike clipping+scaling quantisation techniques, share the pool of weights across the entire network. The work by \cite{Wu2018} is of particular interest since both the motivation and approach are related to ours. Here the authors use a \emph{spectrally relaxed} k-means regularisation term to encourage the network weights to be more amenable to clustering. In their case, they focus on a filter-row codebook inspired by the row-stationary dataflow used in some accelerator designs \cite{Chen2017}. However, their formulation is explored only for convolution, and they restrict clustering to groups of weights (filter rows) rather than individual weights due to computational limitations as recalibrating the k-means regularisation term is expensive during training. 

Similarly, \cite{Stock2019, Fan2020} focus on quantising groups of weights into single codewords rather than the individual weights themselves.  Weight-sharing approaches similar to ours include \cite{Ullrich2017}.  The authors use the distance from an evolving Gaussian mixture as a regularisation term to prepare the clustering weights. Although it is successful with small dataset-model combinations, the complex optimisation --- particularly the additional requirement for Inverse-Gamma priors to lower-bound the mixture variance to prevent mode collapse --- limits the method's practical applicability due to the high computational costs of training.  In our formulation, the weights already fixed no longer contribute to the regularisation prior, reducing the computational overhead.  We further reduce computation by not starting with a complete set of cluster centres but add them iteratively when needed.

\section{Method}

\begin{algorithm} \label{alg:clustering}
\small
\LinesNumbered
\DontPrintSemicolon
 \While{$|W^{t+1}_{\mathrm{fixed}}| \leq Np_t$}{ 
   $\co \leftarrow 0$ \;
   $\mathrm{fixed}_{\mathrm{new}}$ $ \leftarrow [ \ ]$ \;
   \While{$\mathrm{fixed}_{\mathrm{new}}$  \text{is empty}}{
    Increase the order $\co \leftarrow \co  + 1$ \;
        for each $i=1\ldots,|W^{t+1}_{\mathrm{free}}|$ \;
        \, \, $c^\co_*(i)=\min_{c\in\Co} \drel(w_i,c)$ \;
        for each cluster centre $c_k^\co \in \Co$ \;
        \, \, $n^\co_k=\sum_i\ind [c^\co_k=c^\co_*(i)]$  \;
        $k^*=\arg\max_k n_k^\co$ \;
        Sort: $[w'_1, \ldots, w'_{N}] \leftarrow [w_1, \ldots, w_{N}]$, $w'_i=w_{\pi(i)}$, $\pi$ permutation \; \, \, where  $\drel(w'_i, c^\co_{k*})< \drel(w'_{i+1}, c^\co_{k*})$\; 
      
        $i=1$, $\mathrm{mean} = \drel(w'_1, c^\co_{k*})$ \;
        \While{$\mathrm{mean}\leq \delta^t$}{
    $\mathrm{fixed}_{\mathrm{new}}\leftarrow w'_i$ \;
    $\mathrm{mean} \leftarrow \frac{i}{i+1}*\mathrm{mean} + \frac{1}{i+1} \drel(w'_{i + 1}, c^\co_{k*})$ \;
    $i\leftarrow i+1$ \;
   }
    }
     Assign all the weights in $\text{fixed}_{\text{new}}$ to cluster centre $c^\co_*(i)$, moving them from  $W^{t+1}_{\mathrm{free}}$ to $W^{t+1}_{\mathrm{fixed}}$
    } 
 \caption{Clustering $Np_t$ weights at the $t^{th}$ iteration. }
\end{algorithm}

\noindent \textbf{Quantisation.} Consider a network $\mathcal{N}$ parameterised by $N$ weights $W= \{w_1,$ $..., w_N\}$. Quantising a network is the process of reformulating $\mathcal{N}' \leftarrow \mathcal{N}$ where the new network $\mathcal{N}'$ contains weights which all take values from a reduced pool of $k$ cluster centres $C= \{c_1, ..., c_{k}\}$ where $k \ll N$.  After quantisation,  each of the connection weights in the original network is replaced by one of the cluster centres $w_i \leftarrow c_j$, $W' = \{ w'_i | w'_i \in C,  i=1, \cdots, N \},  \  |W'| = k$, where $W'$ is the set of weights of the new network $\mathcal{N}'$,  which has the same topology as the original $\mathcal{N}$.   

\noindent \textbf{Method Outline.} WFN is comprised of $T$ \emph{fixing iterations} where each iteration $t \in T$ has a training and a clustering stage.   The clustering stage is tasked with partitioning the weights into two subsets $W=  W_{\text{fixed}}^t \ \cup \ W_{\text{free}}^t$. $W_{\text{fixed}}^t$ is the set of weights set equal to one of the cluster centre values $c_k \in C$. These \emph{fixed}  weights $w_i \in W^t_{\text{fixed}}$ are not updated by gradient decent in this,  nor any subsequent training stages.   In contrast, the \emph{free-weights} denoted by $W_{\text{free}}^t$ remain trainable during the next training stage. With each subsequent iteration $t$ we increase the proportion   $ p_t = \frac{|W^t_{\text{fixed}}|}{|W|}$ of weights that take on fixed cluster centre values, with $p_0 < p_1 \ldots < p_T=1$. By iteration $T$,  all weights will be fixed to one of the cluster centres.  The training stage combines gradient descent  on a cross-entropy classification loss, along with a regularisation term that encourages tight-clusters, in order to maintain lossless performance as we fix more of the weights to cluster centres.

\noindent \textbf{Clustering Stage.} In the clustering stage, we work backwards from our goal of minimising the relative distance travelled for each of the weights to determine which values cluster centres $c_i \in C$ should take.  For a weight $w_i \in W$ and cluster centre $c_j \in C$ we define a relative distance measure $D_{\text{rel}}(w_i, c_j) =  \frac{|w_i-c_j|}{|w_i|}$.  To use this in determining the cluster centres, we enforce a threshold $\delta$ on this relative distance,  $D_{\text{rel}}(w_i, c_j) \leq \delta$ for small $\delta$.  We can then define the cluster centres  $c_j \in C$ which make this possible using a simple recurrence relation. Assume we have a starting cluster centre value $c_j$,  we want the neighbouring cluster value $c_{j+1}$ to be such that if a network weight $w_i$ is between these clusters  $w_i \in [c_j, \frac{c_{j+1} + c_{j}}{2}]$ then $D_{\text{rel}}(w_i, c_{j}) \leq \delta$.  Plugging in $\frac{c_{j+1} + c_j}{2}$ and $c_j$ into $D_{\text{rel}}$ and setting it equal to $\delta$ we have:
\begin{equation}
\frac{| \frac{c_{j+1} + c_{j}}{2} - c_{j} | }{\frac{c_{j+1} + c_{j}}{2}} =  \delta, \mbox{ leading to }
c_{j+1} =  c_j (\frac{1 + \delta}{1 - \delta}),  \  0 <  \delta < 1,
\end{equation}
\noindent a recurrence relation that provides the next cluster centre value given the previous one.  With this, we can generate all the cluster centres given some boundary condition  $c_0=\delta_0$. $\delta_0$ is the  lower-bound cluster threshold,  and all weights $w_i$ for $|w_i|<\delta_0$ are set to $0$ (pruned).  This lower bound serves two purposes: firstly, it reduces the number of proposal cluster centres which would otherwise increase exponentially in density around zero,  and additionally, the zero-valued weights  makes the network more sparse. This will allow sparsity-leveraging hardware to avoid operations that use these weights,  reducing the computational overhead.  As an upper-bound, we stop the recurrence once a cluster centre is larger than the maximum weight in the network, $\max_{j} |c_j| \leq \max_{i} |w_i|,  \ w_i \in W,  c_j \in C $.  

\begin{figure}
\centering
\includegraphics[width=  \columnwidth]{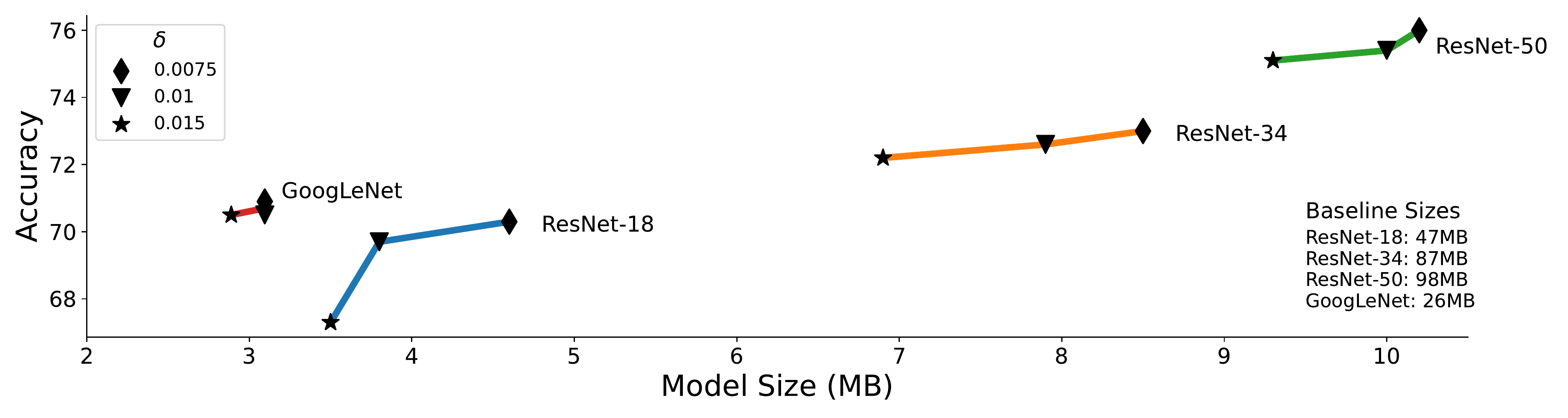}
\caption{The accuracy vs model size trade-off can be controlled by the  $\delta$ parameter.  All experiments shown are using the ImageNet dataset, accuracy refers to top-1.  }
\label{fig:trade-off}
\end{figure}

\noindent \textbf{Generating the Proposed Cluster Centres.} Putting this together, we have a starting point $c_0 = \delta_0$, a recurrence relation to produce cluster centres given $c_0$ that maintains a relative distance change when weights are moved to their nearest cluster centre, and a centre generation stopping condition  $c_j \leq \max_{i \in W} |w_i|, $ $ c_j \in C  $. We use the $\delta_0$ value as our first proposed cluster centre $c_0$ with the recurrence relation generating a proposed cluster set of size $s$. Since all these values will contain only positive values, we join this set with its negated version along with a zero value to create a proposal cluster set $C^S = \{ a (\frac{1+\delta}{1-\delta})^j \delta_0 \ | \ j = 0, 1 \cdots s; \ a= +1,0, -1 \}$ of size $2s + 1$.  

To account for the zero threshold $\delta_0$ and for ease of notation as we advance, we make a slight amendment to the definition of the relative distance function $D_{\text{rel}}(w_i, c_j)$: 
\begin{equation}
D^{+}_{\text{rel}}({w_i, c_j}) = \begin{cases}
\frac{|w_i - c_j|}{|w_i|}, & \text{if} \ |w_i| \geq \delta_0 \\
0 & \text{otherwise.}
\end{cases}
\end{equation}

\noindent \textbf{Reducing $k$ with Additive Powers-of-two Approximations.} Although using all of the values in $C^{S}$ as centres to cluster the network weights would meet the requirement for the relative movement of weights to their closest cluster to be less than $\delta$, it would also require a large number of $k = |C^S|$ clusters. In addition, the values in $C^S$ are also of full 16-bit precision, and we would prefer many of the weights to be powers-of-two for ease of multiplication in hardware. With the motivation of reducing $k$ and encouraging powers-of-two clusters whilst maintaining the relative distance movement where possible,  we look to a many-to-one mapping of the values of $C^{S}$ to further cluster the cluster centres.   Building on the work of others \cite{Zhou2017, YuhangLiXinDong2020}, we map each of the values  $c_i \in C^{S}$ to their nearest power-of-two, $\mathrm{round}(c_i) = \mathrm{sgn}(c_i) 2^{\round{\log_2(c_i)}}$ and, for flexibility,  we further allow for \emph{additive} powers-of-two rounding.  With additive powers-of-two rounding,  each cluster value can also come from the sum of powers-of-two values ($b$-bit) up to order $\co$ where the order represents the number of powers-of-two that can contribute to the approximation.

\noindent \textbf{Minimalist Clustering.} We are now ready to present the clustering procedure for a particular iteration $t$, which we give the pseudo-code for in Algorithm \ref{alg:clustering}.   We start the iteration with $\omega=1$ and a set of weights not yet fixed $W^t_{\text{free}}$.  For the set of cluster centres $\Co$ of order $\omega$, let $c^\co_*(i)=\min_{c\in\Co} \drel(w_i,c)$ be the one closest to weight $w_i$.  $n^\co_k=\sum_i\ind [c^\co_k=c^\co_*(i)]$ counts the number of weights assigned to cluster centre $c_k^\co\in\Co$, where the indicator function $\ind [x]$ is 1 if $x$ is true and 0 otherwise. Let $k^*=\arg\max_k n_k^\co$ so that $c^\co_{k^*}$ is the modal cluster.  For the cluster $k^*$ let permutation $\pi$ of $\{1, \ldots, N\}$ that maps $w_i \mapsto w'_{\pi(i)}$,  be such that the sequence $(w'_1(k^*), w'_2(k^*), \ldots, w'_N(k^*))$ is arranged in ascending order of relative distance from the cluster $c^\co_{k^*}$.  In other words, $\drel(w'_i(k^*), c^\co_{k^*}) \leq \drel(w'_{i+1}(k^*), c^\co_{k^*})$, for $i=1, \ldots, (N-1)$.  We choose $n$ to be the largest integer such that:
\begin{equation}
\sum_{i=1}^n \drel(w'_i(k^*), c^\co_{k^*}) \leq n\delta,  \mbox{ and } \sum_{i=1}^{n+1} \drel(w'_i(k^*), c^\co_{k^*}) > (n+1) \delta,
\end{equation}

\noindent and define $\{w'_1, w'_2, \ldots, w'_n\}$ to be the set of weights to be fixed at this stage of the iteration.  These are the weights that can be moved to the cluster centre $c^\co_{k^*}$ without exceeding the average relative distance $\delta$ of the weights from the centre.  The corresponding weight indices from the original network $\mathcal{N}$ are in $\{\pi^{-1}(1), \ldots, \pi^{-1}(n)\}$, and called $\text{fixed}_{\text{new}}$in the algorithm.   If there are no such weights that can be found, \emph{i.e.}, for some cluster centre $l^*$, the minimum relative distance $\drel(w'_1(l^*), c_{l^*}) > \delta$, the corresponding set $\text{fixed}_{\text{new}}$ is empty.  In this case,  there are no weights that can move to this cluster centre without breaking the $\delta$ constraint and we  increase order $\co \leftarrow \co + 1$ to compute a new $c^\co_{k^*}$,  repeating the process until $|\mathrm{fixed}_{\mathrm{new}}| > 0$.   Once $\mathrm{fixed}_{\mathrm{new}}$ is non-empty, we fix the identified weights $\{w'_1, w'_2, \ldots, w'_n\}$ to their corresponding cluster centre value $c^\co_{k^*}$  and move them into $W_{\text{fixed}}^{t+1}$.  We continue the process of identifying cluster centres and fixing weights to these centres until $|W_{\text{fixed}}^{t+1}| \geq Np_t$,  at which point the iteration $t$ is complete and the training stage of iteration  $t+1$ begins. Our experiments found that a larger $\delta$ has less impact on task performance during early $t$ iterations and so we use a decaying $\delta$ value schedule to maximise compression: $\delta^t = \delta (T - t + 1)$, $t \in T$. We will show later that,  with a small $\delta$,  over 75\% of the weights can be fixed with $\co = 1$ and over 95\% of weights with $\co \leq 2$.   

\noindent \textbf{Training Stage.} Despite the steps taken to minimise the impact of the clustering stage, without retraining, performance would suffer. To negate this, we perform gradient descent to adjust the remaining free weights $W^t_{\text{free}}$. This allows the weights to correct for any loss increase incurred after clustering where training aims to select values $W^t_{\text{free}}$ that minimise the cross entropy loss $\mathcal{L}_{\text{cross-entropy}}$ whilst $W_{\text{fixed}}$ remain unchanged.   

\noindent \textbf{Cosying up to Clusters.} Having the remaining $W^t_{\text{free}}$ weights closer to the cluster centroids $C$ post-training makes clustering less damaging to performance.  We coerce this situation by  adding to the retraining  loss a regularisation term 
\begin{equation} \label{eq:reg_norm}
    \mathcal{L}_{\text{reg}} = \sum_{i \in W_{\text{free}}}^N \sum_j^k D_{\text{reg}}^+(w_i, c_j) p(c_j | w_i),
\end{equation}
 where $p(c_j | w_i) = \frac{e^{-D_{\text{reg}}^+(w_i, c_j)}}{\sum_l^k e^{-D_{\text{reg}}^+(w_i, c_l)}}$. The  idea is to penalise the free-weights $W^t_{\text{free}}$  in proportion to their distance to the closest cluster.  Clusters that are unlikely to be weight $w_i$'s nearest --- and therefore final fixed value ---  do not contribute much to the penalisation term.  We update the gradients of the cross-entropy training loss with the regularisation term:
 \[
 \mathbf{w} \leftarrow \mathbf{w} - \eta\left( \nabla_{\mathbf{w}} \mathcal{L}_{\mathrm{cross-entropy}} +  \alpha\frac{\mathcal{L}_{\mathrm{cross-entropy}}}{\mathcal{L}_{\mathrm{reg}}}
\nabla_{\mathbf{w}}\mathcal{L}_{\mathrm{reg}} \right), 
 \]
with $\alpha$ a hyper-parameter, and $\eta$ the learning rate schedule. In our implementation we name $\alpha$ times the ratio of loss terms as $\gamma$, and we detach $\gamma$ from the computational graph and treat it as a constant. 

\begin{figure}
\centering
\includegraphics[width= 0.9\textwidth]{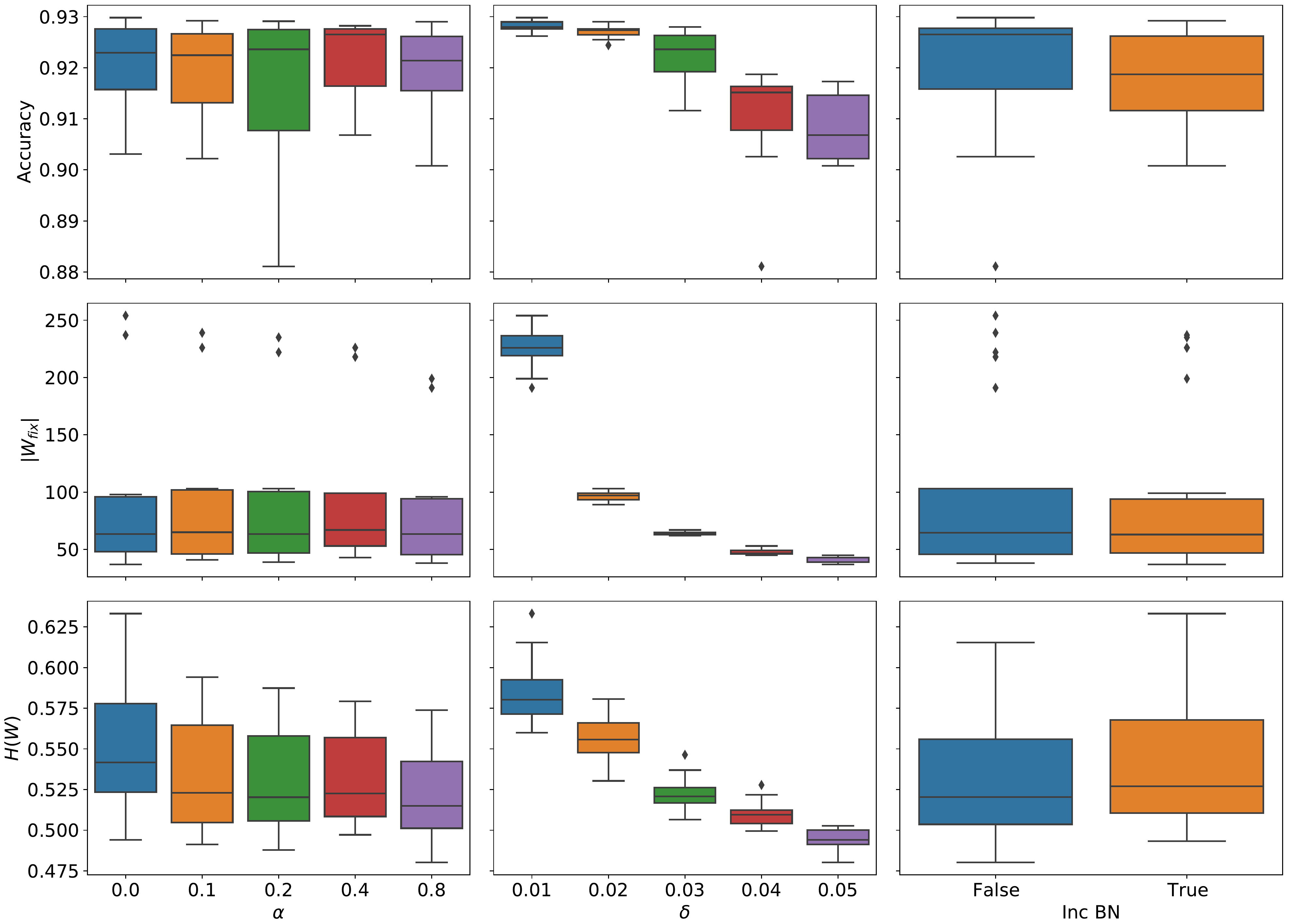}
\caption{Exploring the hyper-parameter space with ResNet18 model trained on the CIFAR-10 dataset. Columns; \textbf{Left:} varying the regularisation ratio $\alpha$,  \textbf{Middle:} varying the distance change value $\delta$,  \textbf{Right:} whether we fix the batch-norm variables or not. Rows; \textbf{Top:} top-1 accuracy test-set CIFAR-10, middle: total number of unique weights, bottom: entropy of the weights. }
\label{fig:hyperparamsearch}
\end{figure}

\section{Experiment Details} 

We apply WFN to fully converged models trained on the CIFAR-10 and ImageNet datasets. Our pre-trained models are all publicly available with strong baseline accuracies\footnote{CIFAR-10 models : https://github.com/kuangliu/pytorch-cifar, ImageNet models: torchvision}: Resnet-(18,34,50) \cite{he2016deep} and, GoogLeNet \cite{Chollet2017a}.  We run ten weight-fixing iterations for three epochs, increasing the percentage of weights fixed until all weights are fixed to a cluster.  In total, we train for 30 epochs per experiment using the Adam optimiser \cite{Kingma2015} with a learning rate $2\times10^{-5}$. We use grid-search to explore hyper-parameter combinations using ResNet-18 models with the CIFAR-10 dataset and find that the regularisation weighting $\alpha  = 0.4$ works well across all experiments reducing the need to further hyper-parameter tuning as we advance.   The distance threshold $\delta$ gives the practitioner control over the compression-performance trade-off (see Figure \ref{fig:trade-off}),  and so we report a range of values.  We present the results of a hyper-parameter ablation study using CIFAR-10 in the Figure \ref{fig:hyperparamsearch}.

\section{Results} 

\begin{figure}[h]
\centering
\includegraphics[width = \columnwidth]{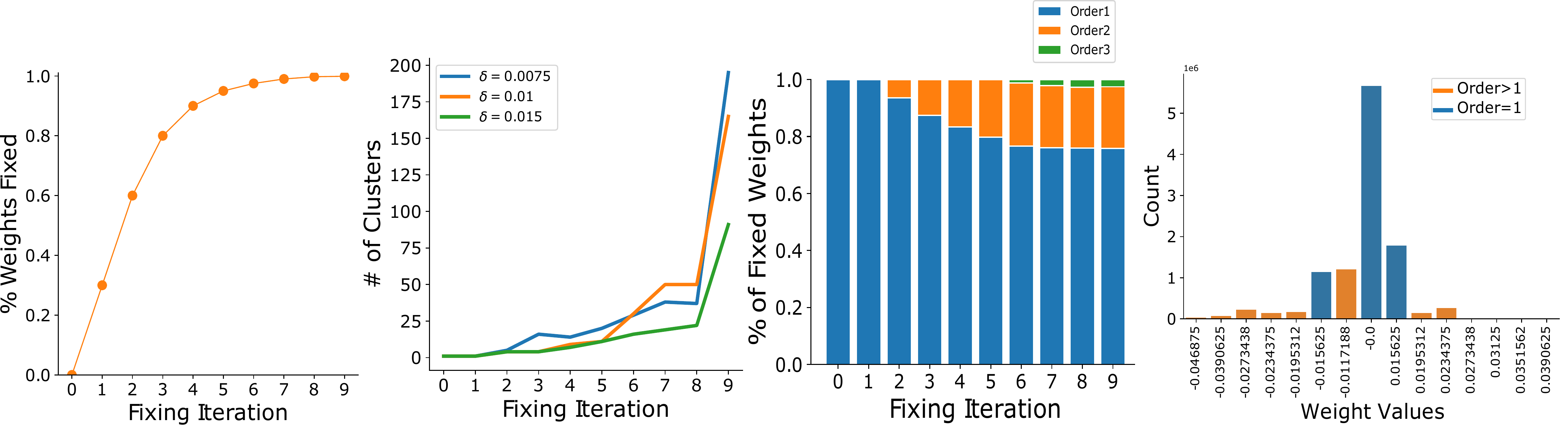}
\caption{\textbf{Far left:} We increase the number of weights in the network that are fixed to cluster centres with each fixing iteration.  \textbf{Middle left:} Decreasing the $\delta$ threshold increases the number of cluster centres,  but only towards the last few fixing iterations, which helps keep the weight-space entropy down.  \textbf{Middle right:} The majority of all weights are order 1 (powers-of-two),  the increase in order is only needed for outlier weights in the final few fixing iterations.   \textbf{Far right:} The weight distribution (top-15 most used show) is concentrated around just four values.  }
\label{fig:weight_dist}
\end{figure}

\noindent We begin by comparing WFN for a range of $\delta$ values against a diverse set of quantisation approaches that have comparable compression ratios (CR) in Table \ref{tab:imnetgen}. The 3-bit quantisation methods we compare include: LSQ \cite{Esser2019}, LQ-Net \cite{Zhang} and APoT \cite{YuhangLiXinDong2020}.   We additionally compare with the clustering-quantisation methods using the GoogLeNet model: Deep-$k$-Means \cite{Wu2018} whose method is similar to ours,  KQ \cite{Yu2020},  and GreBdec \cite{Yu2017}. Whilst the results demonstrate WFN's lossless performance with SOTA CR,  this is not the main motivation for the method. Instead, we are interested in how WFN can reduce the number of unique parameters in a network and corresponding weight-space entropy as well as the network representational cost, as defined in \cite{Wu2018}. This metric has been tested and verified to linearly correlate with energy estimations deduced using the energy-estimation tool proposed in \cite{yang2017designing}: $\mathrm{Rep}(\mathcal{N'}) = |W| N_w B_w$.

\begin{table}[bt!] 
\caption{A comparison of WFN against other quantisation and weight clustering approaches.  The WFN  pipeline is able to achieve higher compression ratios than the works compared whilst matching or improving upon baseline accuracies.  }
\centering
\scriptsize
\begin{threeparttable}
\begin{tabular} {c c  c c  c | c c   c c c }
\hline

& & &   \multicolumn{2}{c}{\textbf{Accuracy (\%)}} & & &     \multicolumn{2}{c}{\textbf{Accuracy (\%)}} &   \\ 
 \textbf{Model} &  \textbf{Method} &    \textbf{Top-1} &  \textbf{Top-5} &  \textbf{CR} &
  \textbf{Model} &  \textbf{Method} &   \textbf{Top-1} &  \textbf{Top-5} &  \textbf{CR} \\
\hline

ResNet-18  & Baseline  & 68.9 &  88.9 & 1.0  & ResNet-34 & Baseline  & 73.3 & 90.9 & 1.0 \\

& LQ-Net & 68.2 & 87.9 & 7.7 & & LQ-Net &  71.9 & 90.2 &  8.6   \\

& APoT   & 69.9 & 89.2 & 10.2  & & APoT    & 73.4 & 91.1 & 10.6  \\

& LSQ   &  70.2\tnote{+}   & 89.4\tnote{+}  & 9.0\tnote{*}  & &  LSQ  & 73.4\tnote{+}  & 91.4\tnote{+}  & 9.2\tnote{*}  \\

& WFN   $\delta= 0.015$ & 67.3 & 87.6  &  13.4 & & WFN  $\delta= 0.015$  & 72.2 &  90.9 & 12.6 \\

& WFN   $\delta= 0.01$  & 69.7 &89.2 & 12.3 & & WFN   $\delta= 0.01$  & 72.6 & 91.0 & 11.1 \\

& WFN   $\delta= 0.0075$  & 70.3 & 89.1 &  10.2 & & WFN   $\delta= 0.0075$  & 73.0 &  91.2 & 10.3 \\
\hline 
ResNet-50   &  Baseline  & 76.1 & 92.8 & 1.0
& GoogLeNet & Baseline  & 69.7 & 89.6 & 1.0

\\

 & LQ-Net  & 74.2 & 91.6 & 5.9 & & Deep $k$-Means  & 69.4 & 89.7 & 3.0 \\

& APoT  & 75.8 & 92.7 & 9.0 & 
  & GreBdec & 67.3 & 88.9 & 4.5 \\

& LSQ & 75.8\tnote{+}  & 92.7\tnote{+} & 8.1\tnote{*} & &  KQ  & 69.2 & - & 5.8 \\

& WFN  $\delta= 0.015$   & 75.1 & 92.1 &  10.3 & & WFN  $\delta= 0.015$  & 70.5 &  89.9 & 9.0 \\

& WFN  $\delta= 0.01$ & 75.4 & 92.5 &  9.8 & & WFN  $\delta= 0.01$  & 70.5 &  90.0  & 8.4 \\

& WFN   $\delta= 0.0075$ & 76.0 & 92.7 &  9.5 & & WFN   $\delta= 0.0075$ & 70.9 &  90.2 & 8.4 \\
\label{tab:imnetgen}

\end{tabular}
\begin{tablenotes}
\item[*] Estimated from the LSQ paper model size comparison graph, we over-estimate to be as fair as possible. 
\item[+] Open-source implementations have so far been unable to replicated the reported results: https://github.com/hustzxd/LSQuantization. 
\end{tablenotes}
\end{threeparttable}
\end{table}


%

\noindent Here, the representation cost is measured as the product of  $N_w$, the number of operations weight $w$ is involved in, $B_w$ its bit-width and $|W|$, the number of unique weights in the network, respectively.

Due to the low weight-space entropy we achieve,  we suggest Huffman encoding to represent the network weights (as is used by various accelerator designs \cite{Moons2016,  Han2016}). Given that the weight-representational bit-width will vary for each weight, we amend the original formulation to account for this, introducing
\begin{equation} \label{eq:repcost}
\mathrm{Rep}_{\mathrm{Mixed}}(\mathcal{N}') = \sum_{w_i \in W} N_{w_i} B_{w_i} 
\end{equation}
\noindent Here $N_{w_i}$ is the number of times $w_i$ is used in an inference computation, and $B_{w_i}$ its Huffman-encoded representation bit-width of $w_i$. 

\begin{table*}[t] \label{tab:resnetim} 
\centering
\scriptsize
\begin{threeparttable}
\caption{A full metric comparison of WFN Vs. APoT. Params refers to the total number of unique parameter values in the network. No BN-FL refers to the unique parameter count not including the first-last and batch-norm layers.   WFN outperforms APoT even when we discount the advantage gained of taking on the challenge of quantising all layers. Model sizes are calculated using LZW compression.  }
\label{tab:imnetbreakdownrep}
\begin{tabular}{c c c c c c c c}
\centering
\textbf{Model} & \textbf{Method} & \textbf{Top-1} & \textbf{Entropy} &  \textbf{Params}  &  \textbf{No BN-FL} & \textbf{$\mathbf{Rep}_{\mathbf{Mixed}}$} &  \textbf{Model Size} \\
\hline 
ResNet-18 & Baseline & 68.9 & 23.3  & 10756029 & 10276369  & 1.000 & 46.8MB   \\
& APoT (3bit)  & 69.9   &  5.77 &   9237 & 274  & 0.283 & 4.56MB  \\
& WFN   $\delta= 0.015$ & 67.3  &
 2.72 &  90 & 81 & 0.005 & 3.5MB  \\
 & WFN   $\delta= 0.01$ & 69.7  &
 3.01 &  164 & 153 & 0.007 & 3.8MB   \\
& WFN   $\delta= 0.0075$ & 70.3 & 4.15 &  193 & 176 & 0.018 & 4.6MB  \\
\hline

ResNet-34 & Baseline & 73.3  & 24.1  & 19014310 & 18551634 & 1.000 & 87.4MB  \\
& APoT (3bit)  & 73.4   &  6.77 &  16748 & 389  & 0.296 & 8.23MB    \\
& WFN   $\delta= 0.015$ & 72.2  &
2.83 & 117 & 100 & 0.002 &  6.9MB    \\
 & WFN   $\delta= 0.01$ & 72.6   &
 3.48 & 164 & 130 & 0.002 &  7.9MB \\
& WFN   $\delta= 0.0075$ & 73.0   & 3.87 &  233 & 187 & 0.004 & 8.5MB   \\
\hline
ResNet-50* & Baseline & 76.1  & 24.2  & 19915744 & 18255490 & 1.000 & 97.5MB  \\
& WFN   $\delta= 0.015$ & 75.1 &
3.55 & 125 & 102 & 0.002 &   9.3MB    \\
 & WFN   $\delta= 0.01$ & 75.4   &
 4.00 & 199 & 163 &  0.002 & 10.0MB  \\
& WFN   $\delta= 0.0075$ & 76.0   & 4.11  &  261 & 217 & 0.003 & 10.2MB    \\
\hline
 \end{tabular}
\begin{tablenotes}
\item[*] The APoT model weights for ResNet-50 were not released so we are unable to conduct a comparison for this setting.
\end{tablenotes}
\end{threeparttable}

\end{table*}

The authors of the APoT have released the quantised model weights and code. We use the released model-saves\footnote{https://github.com/yhhhli/APoT\_Quantization} of this SOTA model to compare the entropy,  representational cost, unique parameter count,  model size and accuracy in Table \ref{tab:imnetbreakdownrep}.    Our work outperforms APoT in weight-space entropy,  unique parameter count and weight representational cost by a large margin. Taking the ResNet-18 experiments as an example,  the reduction to just 164 weights compared with APoT's 9237 demonstrates the effectiveness of WFN. This huge reduction is partly due to our full-network quantisation (APoT, as aforementioned,  does not quantise the first, last and batch-norm parameters). However, this does not tell the full story; even when we discount these advantages and look at weight subsets ignoring the first,  last and batch-norm layers, WFN uses many times fewer parameters and half the weight-space entropy --- see column `No BN-FL' in Table \ref{tab:imnetbreakdownrep}.    Finally, we examine how WFN achieves the reduced weight-space entropy in Figure \ref{fig:weight_dist}. Here we see that not only do WFN networks have very few unique weights,  but we also observe that the vast majority of all of the weights are a small handful of powers-of-two values (order 1). The other unique weights (outside the top 4) are low frequency and added only in the final fixing iterations.

\section{Conclusion}

We have presented WFN,  a pipeline that can successfully compress whole neural networks. The WFN process produces hardware-friendly representations of networks using just a few unique weights without performance degradation. Our method couples a single network codebook with a focus on reducing the entropy of the weight-space along with the total number of unique weights in the network. The motivation is that this combination of outcomes will offer accelerator designers more scope for weight re-use and the ability to keep most/all weights close to computation to reduce the energy-hungry data movement costs. Additionally, we believe our findings and method offer avenues of further research in understanding the interaction between network compressibility and generalisation, particularly when viewing deep learning through the minimal description length principle lens. \\

\noindent \textbf{Acknowledgements.} This work was supported by the UK Research and Innovation Centre for Doctoral Training in Machine Intelligence for Nano-electronic Devices and Systems [EP/S024298/1]. Thank you to Sulaiman Sadiq for insightful discussions.

\clearpage

\bibliographystyle{splncs}
\bibliography{egbib}

\end{document}


\newcommand{\ind}{\mathbb{I}}

\newcommand{\drel}{D^+_{\mathrm{rel}}}

\newcommand{\co}{\omega}

\newcommand{\Co}{\widetilde{C}^\omega}
\pagestyle{headings}
\mainmatter
\def\ECCV16SubNumber{5083}  

\title{Weight Fixing Networks - Additional Information}

\titlerunning{Weight Fixing Networks}

\authorrunning{Subia-Waud et al.}

\author{Christopher Subia-Waud \& Srinandan Dasmahapatra}
\institute{University of Southampton, Southampton, SO17 1BJ \\ \email{\{cc2u18, sd\}@soton.ac.uk}}

\maketitle

\section{Details of the Powers-of-two Approximation Algorithm}

\begin{figure}
\centering
\includegraphics[width=\columnwidth]{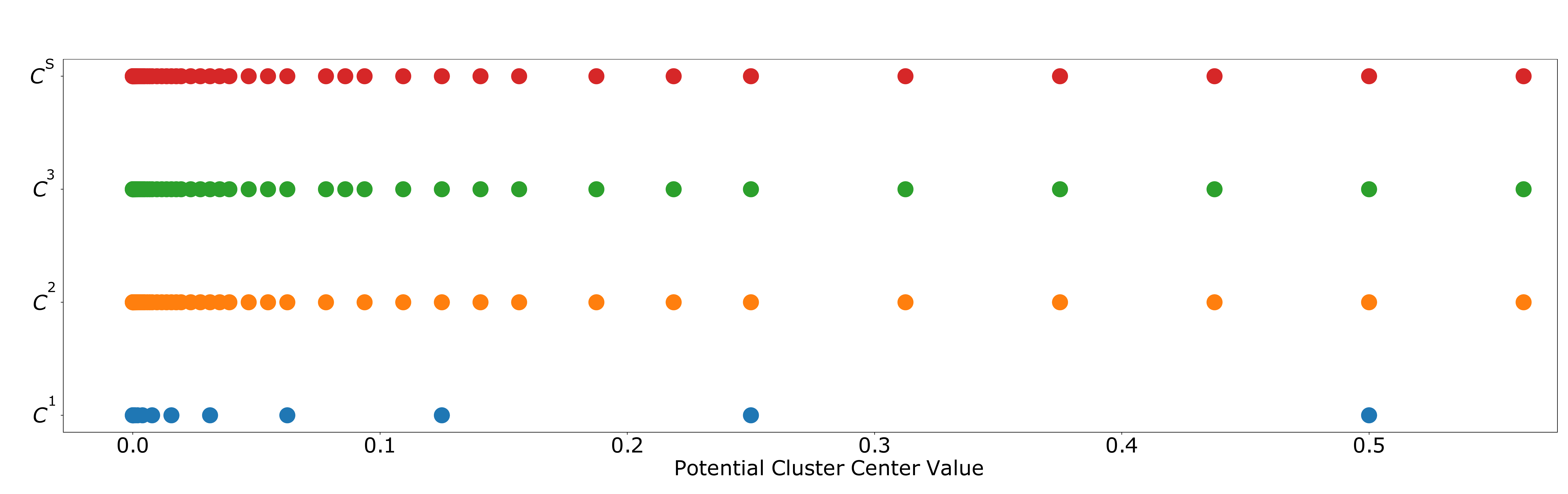}
\caption{Approximating clusters in $C^S$ with different orders}
\label{fig:pow2est}
\end{figure}

We map our proposal set $C^S$ to a $\co$-order approximation where each of the clusters $c_k \in C^S$ are written as $\co$ powers-of-two (Eq \ref{eq:po2}). We do so using  Algorithm \ref{alg:whichweights}.  Figure \ref{fig:pow2est} demonstrates how the values of $C^{S}$ are rounded given different orders.

\begin{equation} \label{eq:po2}
c_k = \sum^{\co}_{j=1} r_j, \ r_j \in \{-\frac{1}{2^b}, \ldots, -\frac{1}{2^{j+1}}, -\frac{1}{2^j}, 0, \frac{1}{2^j}, \frac{1}{2^{j+1}}, \ldots \frac{1}{2^b}\}
\end{equation}

 \begin{algorithm} \label{alg:whichweights}
\DontPrintSemicolon
\LinesNumbered
\SetAlgoLined
\textbf{Input:} The full precision proposal set: $C^S$,  allowable relative distance: $\delta$,  pow2 rounding function: $round(x) = sgn(x)2^{\round{\log_2{(x)}}}$  \;
\textbf{Output:} An order $\co$ precision cluster set: $\Co$  \;
$\Co \leftarrow [ \ ]$ \;
\For{$c_k \in C^S$}{
$c_k' = round(c_k)$ \;
\For{$i = 0 \to \co$}
{
$\delta_{c_k} \leftarrow c_k - c_k' $ \;
\If{$|\delta_{c_k} |   \geq  \delta c_k $}{
$c_k' \leftarrow c_k' + round(\delta_{c_k})$ \;
}}
$\Co \leftarrow  \Co \cup \{c_k'\}$ \;
}
\caption{Determining possible clusters}
\end{algorithm}

\section{Experiment Details} 

We give a full breakdown of the parameters used across all experiments ran in Table \ref{tab:experiments_ran}.

\begin{table}
\centering
\tiny
\begin{tabular}{c c c c c c c c}
Model & Data & Opt & LR & $T$  & Batch size & $\gamma$ & $\alpha$  \\
\hline
ResNet-18 & ImageNet & Adam & 2e-4 & 10 & 128  & \{0.05, 0.025,  ,   0.015, 0.01,  0.0075, 0.005\}  & \{0.2, 0.4\} \\ 
ResNet-34 & ImageNet & Adam & 2e-4 & 10 & 64 & \{0.05, 0.025,  ,   0.015, 0.01,  0.0075, 0.005\} & \{0.4\} \\ 
ResNet-50 & ImageNet & Adam & 2e-4 & 10  & 64 & \{0.05, 0.025,  ,   0.015, 0.01,  0.0075, 0.005\}  & \{0.4\} \\ 
GoogLeNet & ImageNet & Adam & 2e-4 & 10 & 64 & \{0.01,  0.0075, 0.015\} & \{0.4\} \\
ResNet-18 & CIFAR-10 & Adam & 3e-4 & 10  & 512 & \{0.01, 0.02, 0.03, 0.04, 0.05\} & \{0.0, 0.1, 0.2, 0.4, 0.8\}  \\
MobileNet & CIFAR-10 & Adam & 2e-4 & 10  & 512  & \{0.01, 0.02, 0.03, 0.04, 0.05\} & \{0.0, 0.1, 0.2, 0.4, 0.8\}  \\
\end{tabular}
\caption{Full set of hyper-parameters explored for each model-dataset combination.}
\label{tab:experiments_ran}
\end{table}




\section{Additional Analysis}
\subsection{Layerwise Breakdown}

\begin{figure}
\centering
\includegraphics[width=\textwidth]{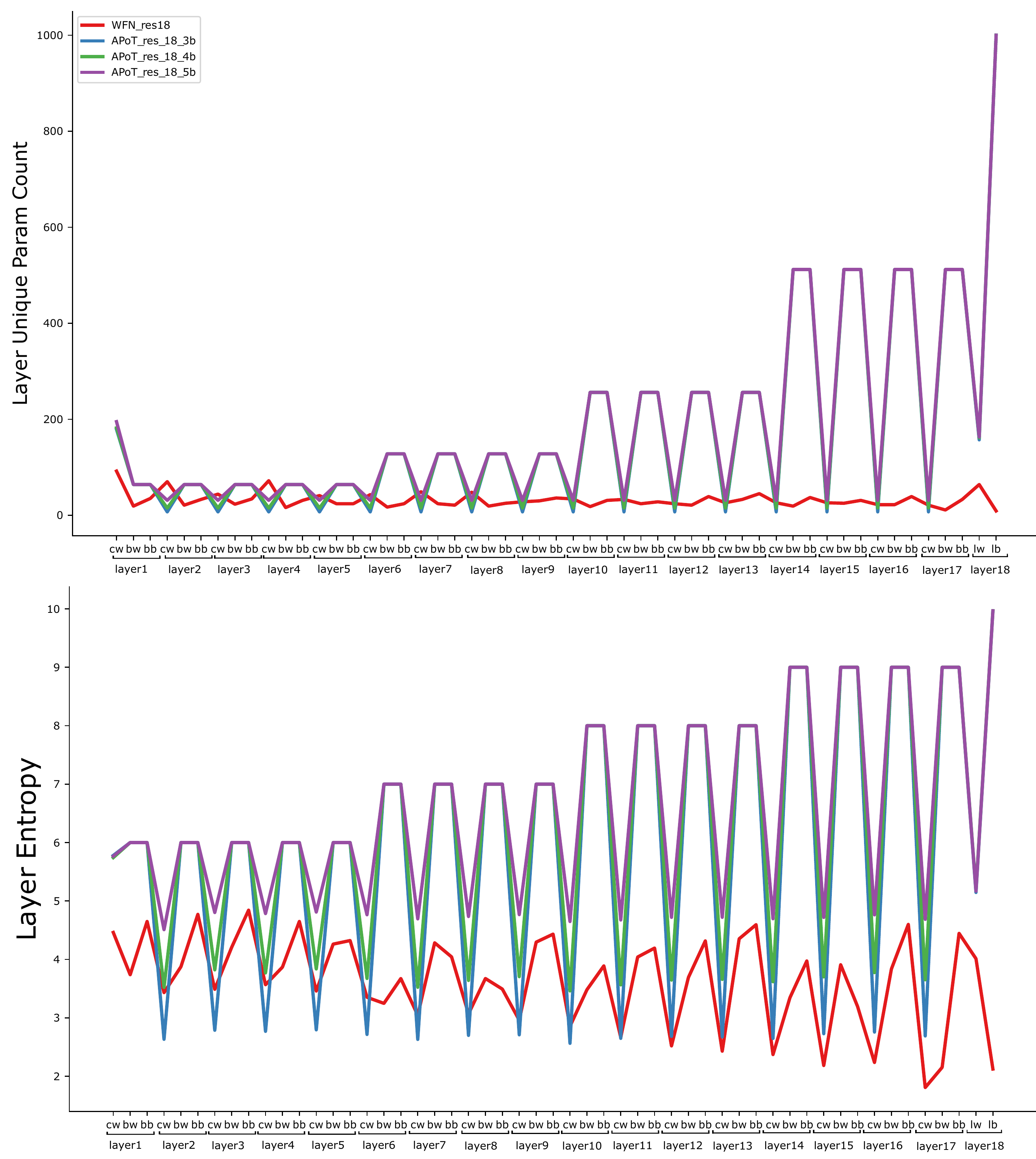}
\caption{We compare WFN with a traditional quantisation set-up (APoT) with varying bit-widths applied to a ResNet18 model trained on the ImageNet dataset. The top chart shows the layerwise unique parameter count where WFN has consistently fewer unique parameters per layer. 
 }
  \label{fig:layerwise}
\end{figure}

In Figure \ref{fig:layerwise} we examine how the parameter count and layer-parameter entropy change with each layer for both the WFN and APoT approaches.  We find both gains over the unquantised layers of APoT, but also that the entropy and parameter count in the convolutional layers (those quantised by APoT) are similar. 

\subsection{A Full Metric Comparison}

In Table \ref{tab:imnetbreakdown} we give the full metric breakdown comparing WFN to the state-of-the-art APoT work.  We calculate the unique parameter count and entropy values for subsets of the weights.  No BN corresponds to all weights other than those in the batch-norm layers, and No BN-FL is the set of weights not including the first-last and batch-norm layers.  It's clear here that WFN outperforms APoT even when we discount the advantage gained of taking on the challenge of quantising all layers. 

\begin{figure}
\centering
\includegraphics[width=\textwidth]{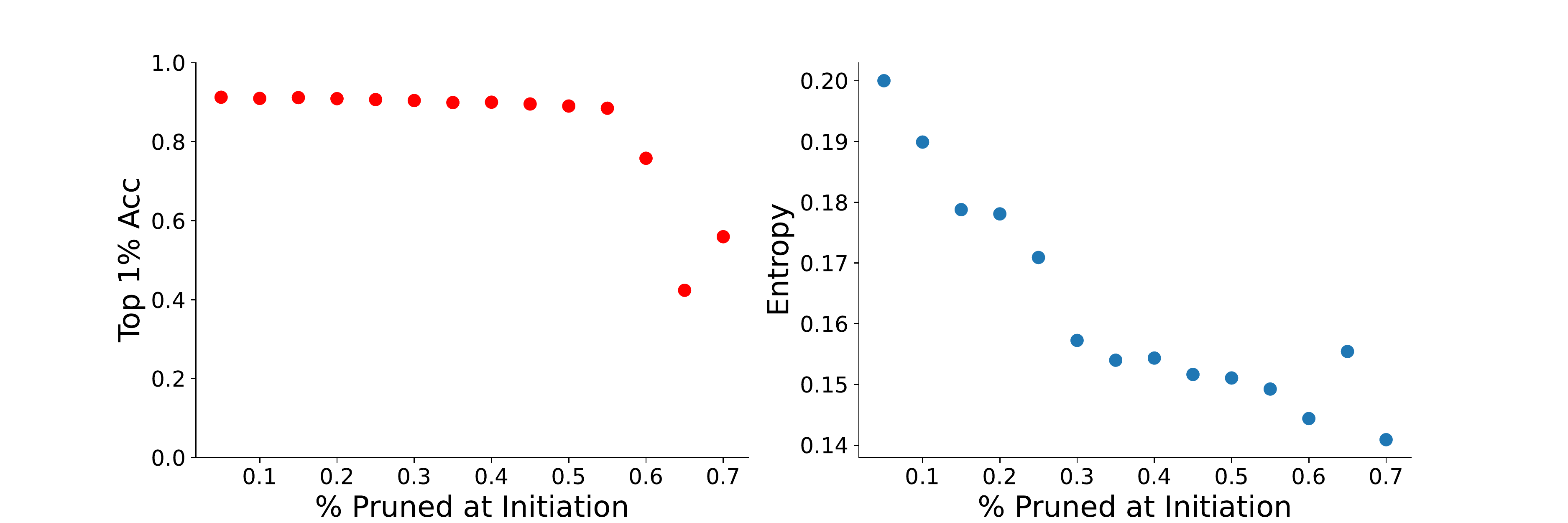}
\caption{Here we show that unstructured pruning at initialisation up to 50\% can be coupled with the WFN process without degradation of performance and can further reduce the weight-space entropy. 
 }
 \label{fig:pruning}
\end{figure}

\begin{table*}[t] 
\centering
\tiny
\begin{threeparttable}
\begin{tabular}{c c  c  c  c  c  c c c}
& &  &   \multicolumn{2}{c}{\textbf{Full Network}} 
&  \multicolumn{2}{c}{\textbf{No BN}}
&  \multicolumn{2}{c}{\textbf{No BN-FL}}  &   \\
\textbf{Model} & \textbf{Method} & \textbf{Top-1} & \textbf{Entropy} &  \textbf{Param Count}  &   \textbf{Entropy} &  \textbf{Param Count} &   \textbf{Entropy} &  \textbf{Param Count}  \\
\hline 
ResNet-18 & Baseline & 68.9 & 23.3  & 10756029 & 23.3  & 10748288 & 23.3  & 10276369   \\
& APoT (3bit)  & 69.9   &  5.77 &  9237 & 5.76 & 1430 & 5.47  & 274  \\
& WFN   $\delta= 0.015$)& 67.3  &
 2.72 &  90 &  2.71  & 81 & 2.5  & 81  \\
 & WFN   $\delta= 0.01$) & 69.7  &
 3.01 &  164 &  3.00 &153 & 2.75 & 142  \\
& WFN   $\delta= 0.0075$) & 70.3 & 4.15 &  193 &  4.13 & 176 & 3.98 & 162  \\
\hline
ResNet-34 & Baseline & 73.3  & 24.1  & 19014310 & 24.1 & 18999320 & 24.10  & 18551634   \\
& APoT (3bit)  & 73.4   &  6.77 &  16748  & 6.75  & 16474 & 6.62  & 389    \\
& WFN   $\delta= 0.015$)& 72.2  &
2.83 & 117  &  2.81 & 100 & 2.68   & 100    \\
 & WFN   $\delta= 0.01$) & 72.6   &
 3.48 & 164 & 3.47   & 132 & 3.35 & 130  \\
& WFN   $\delta= 0.0075$) & 73.0   & 3.87 &  233 & 3.85 & 191 & 3.74   & 187   \\
\hline
ResNet-50 & Baseline & 76.1  & 24.2  & 19915744 & 24.2 & 19872598 & 24.20  & 18255490   \\
& WFN   $\delta= 0.015$)& 75.1 &
3.55 & 125  &  3.50 & 105 & 3.42   & 102   \\
 & WFN   $\delta= 0.01$) & 75.4   &
 4.00 & 199 & 3.97   & 169 & 3.88 & 163   \\
& WFN   $\delta= 0.0075$) & 76.0   & 4.11  &  261 & 4.09 & 223 & 4.00  & 217   \\
\hline
\end{tabular}
\end{threeparttable}
\caption{A full metric comparison of WFN Vs. APoT. We compare the unique parameter count and entropy of all parameters in the full network, as well as the same measures but not including the batch-norm layers (No BN)  and the parameters not including the batch-norm and first and last layers (No BN-FL).      }
\label{tab:imnetbreakdown}
\end{table*}

\subsection{Pruning Experiments}

To explore how WFN interacts with pruning we conduct a simple set of experiments.  Instead of starting the WFN process with all weights un-fixed we randomly select $p\%$ of the weights to be pruned in each layer.  We then run WFN as before starting with $p_t = p$,  reducing the number of $T$ iterations.  The results,  shown in Figure \ref{fig:pruning},  are conducted with a ResNet-18 and Cifar-10 combination, painting a mixed picture.  On the one hand,  WFN and pruning at lower levels ($< 50\%$) is well tolerated and provide two benefits, a lower weight-space entropy and few weight-fixing iterations.  On the other hand,  full-precision networks can tolerate much higher ranges of pruning so there it would seem that a certain amount of synergy between the two approaches is present but this is tempered compared to full precision networks.  \\

\noindent It's important to note that WFN already has a form of pruning built-in with the $\delta_0$ value balancing the emphasis on pruning over quantisation.